\documentclass[letterpaper]{article}

\usepackage{natbib,alifeconf}  
\usepackage{hyperref}

\title{DNN Architecture for High Performance Prediction on Natural Videos Loses Submodule's Ability to Learn Discrete-World Dataset}
\author{Lana Sinapayen$^{1,*}$, Atsushi Noda$^{1,*}$ \\
\mbox{}\\
$^1$Sony Computer Science Laboratories, Tokyo, Japan. *These authors contributed equally to the paper.\\
lana.sinapayen@gmail.com} 
\begin{document}
\maketitle

\begin{abstract}
Is cognition a collection of loosely connected functions tuned to different tasks, or can there be a general learning algorithm? If such an hypothetical general algorithm did exist, tuned to our world, could it adapt seamlessly to a world with different laws of nature?  We consider the theory that predictive coding is such a general rule, and falsify it for one specific neural architecture known for high-performance predictions on natural videos and replication of human visual illusions: PredNet. 
Our results show that PredNet's high performance generalizes without retraining on a completely different natural video dataset. Yet PredNet cannot be trained to reach even mediocre accuracy on an artificial video dataset created with the rules of the Game of Life (GoL). We also find that a submodule of PredNet, a Convolutional Neural Network trained alone, reaches perfect accuracy on the GoL while being mediocre for natural videos, showing that PredNet's architecture itself is responsible for both the high performance on natural videos and the loss of performance on the GoL.
Just as humans cannot predict the dynamics of the GoL, our results suggest that there might be a trade-off between high performance on sensory inputs with different sets of rules.


\end{abstract}


\section{Introduction}

In a world where many tasks have been automated to quasi-perfection, the next big goal for Artificial Intelligence (AI) is Artificial General Intelligence (AGI) (\cite{wang2007}). Beyond domain-specific automation, AGI is often defined as AI with human-level performance, able to generalize its knowledge across different domains~(\cite{adams2012, wang2007}). 
Candidate algorithms to realize AGI vary with the evolution of the field~(\cite{goertzel2010, goertzel2014}), and recent theories include a combination of Deep Learning Networks to emulate human cortical networks~(\cite{yamakawa2017}), neural computation at the edge of chaos~(\cite{smith2016deep, cocchi2017criticality, garson1996cognition}), and of course different implementations of predictive coding~(\cite{hawkins2007,friston2010free,van2010minimum}).

Here we are interested in predictive coding: the idea that brains generate models of the world by learning to predict their own sensory inputs. Although the specific mechanism is up to debate~(\cite{friston2009, garalevicius2007memory}), predictive coding has been found to take place in the human nervous system~(\cite{baldeweg2006repetition, hosoya2005dynamic, rao1999predictive}), especially for visual and auditory processing. In many experiments, specific sensory illusions such as the auditory oddball effect are interpreted as the hallmark of predictive coding~(\cite{schindel2011oddball}). 

A recently proposed Deep Learning architecture, PredNet~(\cite{lotter2016}), has not only demonstrated high performance on natural video prediction, but has been shown to be susceptible to some of the same visual illusions as human beings~(\cite{watanabe2018illusory}) as a side effect of its predictive abilities, despite contemporary explanations of these illusions not relying on predictive coding mechanisms.
No one claims that PredNet accurately recreates the brain processes leading to predictive coding, but PredNed is currently the state of the art architecture replicating both the main consequences and side effects of predictive coding in human vision. 
Can PredNet be used as a general prediction machine, even in visual worlds that humans cannot predict, or is the algorithm's implementation so attuned to human visual processing that it fails in the same circumstances as humans do? In short, does high performance on natural datasets cause a loss in generality for this specific architecture? 

To test this hypothesis, we choose a visual task where simple local spatial rules lead to dynamics that are hard to predict for human beings: the Game of Life automaton (GoL)~(\cite{gardener1970mathematical,Izhikevich2015}). At each time step, the GoL updates its cells depending only on the state of neighboring cells at the previous timestep. It is simple enough to be predicted with accuracy by a simple Convolutional Neural Network~(\cite{rappGOL}). Some dynamics are discrete and some have the feeling of continuous dynamics, for example the ``glider": a bird-like pattern that moves in a constant direction. The produced patterns can be hard to predict; even knowing the underlying rules, humans make mistakes in prediction. We can reasonably extrapolate that prediction without knowing these rules would be even harder for humans.
The main differences between a natural video such as those in the KITTI dataset~(\cite{geiger2013IJRR}) and a GoL-generated video are as follows:
\begin{itemize}
    \item Theoretically, the GoL requires only 1 time step of memory for perfect prediction. Natural videos can show objects with speed or acceleration, which require several time steps to deduce.
    \item
    Natural videos have spatio-temporal continuity: objects maintain their shapes and there is no drastic change from one frame to the other. In the GoL, there is no continuity from one frame to the other and each cell is separate from the others. The GoL is a discrete world.
    \item
    Rules are extremely local in the GoL: a cell's next state only depends on its immediate neighbors.
\end{itemize}


If PredNet had succeeded at this task, it would suggest that the implementation of PredNet captures something fundamental beyond natural rules, and the idea of predictive coding as a general cross-domain learning rule would be strengthened. But our results show that PredNet is attuned to human performance even in its failure modes, as its sensitivity to visual illusions suggested. It comforts this architecture in its place as a replicator of human visual performance, but it also means that the improved performance on natural tasks comes to the cost of performance on tasks that can be solved by  simpler networks. These results stack two different learning architectures (the human brain and PredNet) against predictive coding as a general cross-domain learning rule.

\section{Methods}

All the source code used in this paper is available at~\url{https://github.com/LanaSina/prednet_gol}. The datasets are available at~\url{https://figshare.com/projects/PredNet_Game_of_Life/60971}.

\subsection{PredNet}

\begin{figure}[t]
    \includegraphics[width=\linewidth]{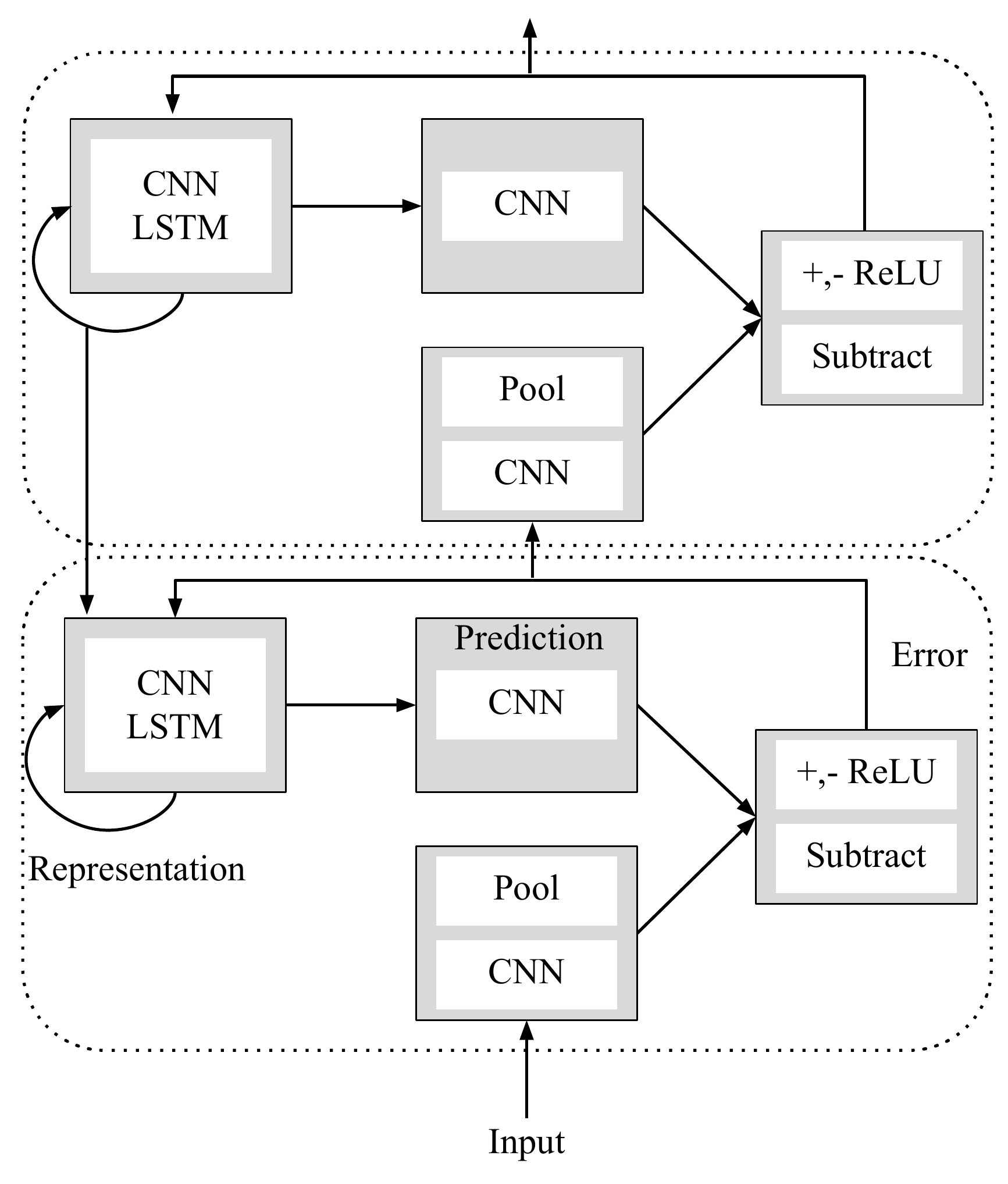}
    \caption{{\bf The architecture of PredNet.} Figure modified from \cite{lotter2016}. The network is divided in modules (solid lines) that are stacked into layers (dashed lines). The input to the first layer is the original image frame at t; the error between the prediction at t and the frame at t+1 is passed as input to the next layer. There are 3 layers in total (only 2 represented here.)
    }
    \label{fig:prednet_structure}
\end{figure}

PredNet is a Deep Learning Neural Network with an architecture based on the principles of predictive coding (Fig.~\ref{fig:prednet_structure}). The network is made of hierarchical layers. Each layer contains several modules, respectively for input, representation, prediction, and error calculation. The input to the first layer is the original image frame at time $t$; the error between the prediction at $t$ and the frame at $t+1$ is passed as input to the next layer, which must therefore learn to predict the error signal from the layer below it. The representations from the upper layer are sent as feedback to the lower layer. 

For our experiments, we use the code provided by~\cite{lotter2016} at this address: \url{https://github.com/coxlab/prednet}. The network has 3 layers, its convolutional modules have filters of size 3 $\times$ 3 pixels.
For the experiment \textbf{without retraining}, we use the weights of the PredNet trained on the KITTI dataset (weights dowloaded directly from their open source code), and test it on different datasets to evaluate the transfer of performance between the different datasets.

For the experiments \textbf{with retraining}, we train PredNet on other datasets and evaluate it on the same datasets as it was trained on.

\subsection{Simple CNN}

 We used a simple CNN Auto-Encoder a minimum model that can learn the GoL rules. The model is similar to the CNN in the Prediction module in the PredNet architecture and uses the same filter size. The architecture is as follows: the encoder consists of one convolutional layer and maps pixels to a latent representation. The filter size is $3 \times 3$ and the stride is set at 1. Zero padding is done to deal with boundary conditions of the image. The decoder uses the transposed architecture. The activation function is ReLU. 
 We implemented this model using Chainer~(\cite{tokui2015chainer}) and trained it with the negative log-likelihood function of the Bernoulli distribution.
 In this model, the data at the episode boundaries is excluded from the training and evaluation data. The number of epochs $\times$ the size of training data is slightly less than one fifth of that of Prednet's KITTI learning.
 
\subsection{Datasets}

We use two existing datasets composed of videos from the real world (called ``natural datasets" in this paper) and one dataset of videos generated from the GoL (called ``artificial dataset").

\subsubsection{- Karlsruhe Institute of Technology and Toyota Technological Institute (KITTI) Dataset}

The KITTI dataset is composed of videos filmed from a car driving 
on a road~(\cite{geiger2013IJRR}). The videos show traffic and pedestrians from the point of view of the driving car. We use the simple raw video part of the dataset (20 fps). As in the original PredNet paper, frames of the videos are downsampled to 128 $\times$ 160 pixels. We used the processing script for outputting frames from videos made available by~(\cite{lotter2016}). The script is also available in the repository published for this paper.

\subsubsection{- First Person Social Interaction (FPSI) Dataset}

The FPSI dataset~(\cite{fathi2012social}) is composed of videos filmed from a first-person point of view. The videos were obtained by fixing cameras to the head of several people who spent a day at a theme park. The original dataset's url~\url{http://cpl.cc.gatech.edu/projects/FPSI/} had become unresponsive at the time of the writing of this paper, so we used the video made available by~\cite{watanabe2018illusory} at \url{https://figshare.com/articles/Training_data/5483668/1}.
We also used their script for extracting and resizing images to 128x160 pixels, but found that the original video was encoded with the wrong FPS parameter. The corrected script is available at our published repository (output: 15 fps).

\subsubsection{- Game of Life (GoL) Dataset}
 
We use Conway's Game of Life (GoL)~(\cite{gardener1970mathematical,Izhikevich2015}) as artificial dataset to test PredNet.
The GoL is a 0-player game where an initial state, decided by the programmer, automatically changes every timestep according to rules.
The GoL is played on a 2D grid where each cell is in one of two possible states: {\it alive} or {\it dead} (white cells are considered {\it alive}, and black cells are considered {\it dead}).
The flow of time is discrete, and the state at the next timestep is completely determined by the state at the previous timestep.
Every cell has 8 neighboring cells (4 cells adjacent orthogonally and 4 cells adjacent diagonally).
The state of a cell depends on how many neighboring cells are {\it alive}. If there are exactly 2 {\it alive} cells in the neighborhood, the state of the central cell does not change. If there are exactly 3 {\it alive} cells in the neighborhood, the next state of the cell is {\it alive} regardless of the current state. For any other number of living cells in in the neighborhood, the next state of the cell is {\it dead} regardless of the current state.
These are the only rules.
Note that all above rules are applied simultaneously.
The interest of the GoL is that simple rules lead to complex patterns. We will briefly introduce the "glider" as an example used in experiments described later.
The glider is a moving pattern composed of 5 cells.
The movement of the glider has four stages; after 4 time steps, the glider has shifted one cell in the diagonal direction.
In the GoL, there are patterns that move while keeping their shape like the glider, patterns that periodically repeat states, and so on.

We generate the video dataset as below.
When generating an initial state, about 10\% of the total cells are randomly generated so as to be {\it alive}. Each pixel represents a cell in the generated 128 $\times$ 160 images.
There are no boundary conditions on the board.
In the GoL, the state often becomes fixed as time passes, so we generate a new initial state every 10 steps. We repeated it 1000 times and used 10,000 images as training data.

\section{Results}

\subsection{Pretrained PredNet performance on KITTI, FPSI and GoL}

We investigate how well the performance of PredNet trained exclusively on the KITTI dataset transfers to other datasets without retraining. We evaluate this model on the original KITTI dataset, on the natural video dataset FPSI, on random GoL patterns, and finally, on a GoL glider.
Note that, in only in this experiment (examining the performance of Pretrained PredNet), when we generate the GoL frames, we first generate 16$\times$20 images and scale up each cell vertically and horizontally by 8 times. Otherwise PredNet did not output anything.

\begin{figure}[t]
    \includegraphics[width=\linewidth]{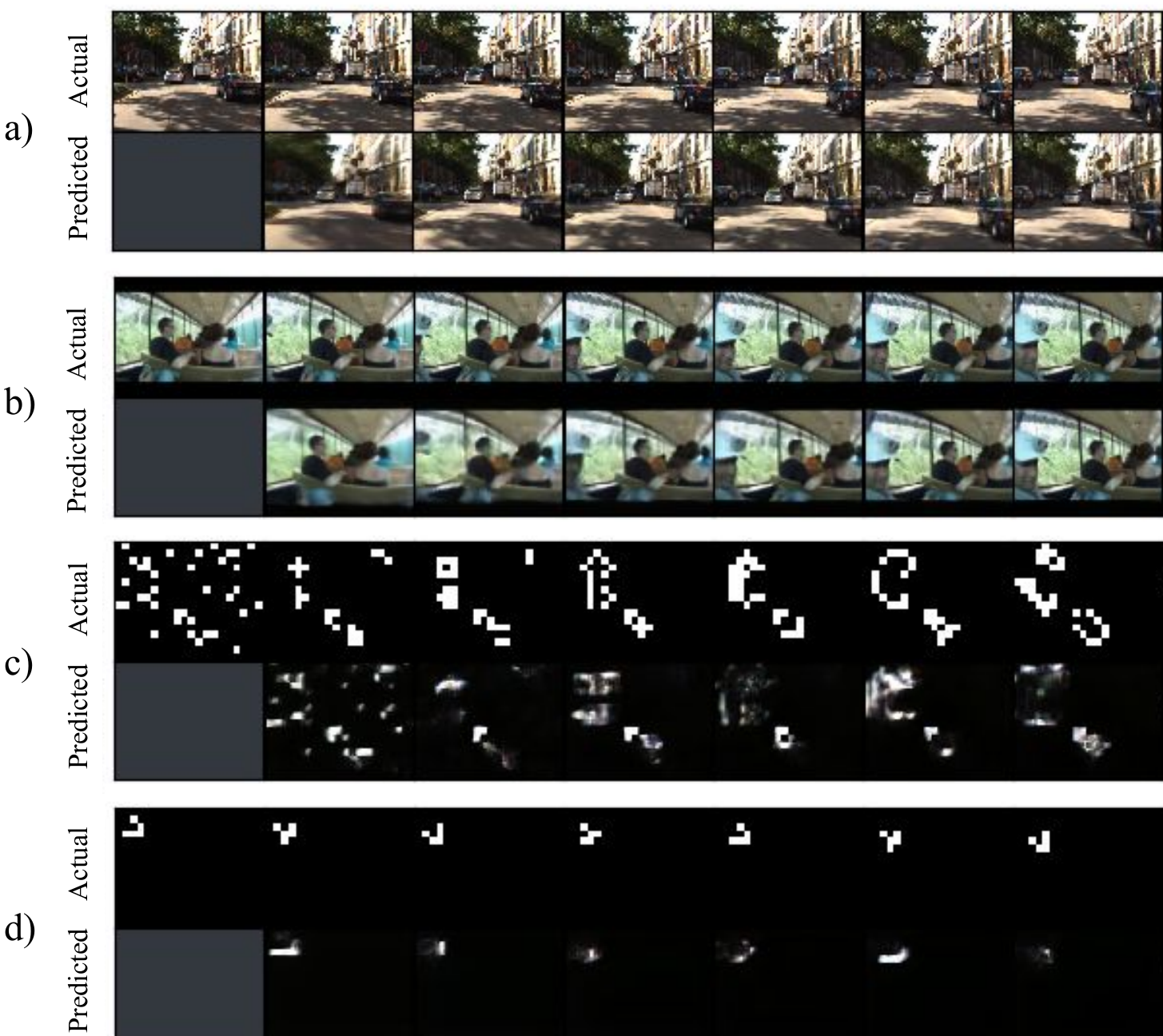}
    \caption{{\bf Predictions generated by a PredNet trained on the KITTI dataset.} See Table~\ref{tab:mse} for measured performance. (a) Predictions for KITTI, the original training dataset. The performance is high. (b) Predictions for the FPSI dataset. The performance is also high. (c) Predictions for a random initial value of the GoL. The performance is poor, as expected. (d) Predictions for a GoL glider. The performance is also poor.}
    \label{fig:pretrained_prednet}
\end{figure}

As shown on Fig.~\ref{fig:pretrained_prednet}-a), b), and Table~\ref{tab:mse}, the performance transfers well between natural datasets. PredNet trained on the KITTI dataset still has better performance on the FPSI dataset than a simple "copy the last frame" model. This result is a testament to the robustness of the PredNet architecture.
On the other hand, as shown on Fig.~\ref{fig:pretrained_prednet}-c) and d), the performance does not transfer to datasets with artificial rules. As expected, the pretrained PredNet cannot predict the GoL, or even gliders. Even if gliders have a partly translational motion, the dynamics of the GoL are too far from the simple translations and rotations that might be responsible for the majority of variations in natural videos.
In addition, we see an interesting error on frame 3 at Fig.~\ref{fig:pretrained_prednet}-b): the translational motion of the cap that appeared at frame 2 is extrapolated, leading to the prediction of a floating cap at frame 3. 
This model cannot know that the cap should be on a human head, even if the person wearing the cap can be seen a few frames earlier. This would require either excellent use of short term memory, existence of an internal world model, or high level inference.

\subsection{Retrained PredNnet performance on the GoL}

Since the performance of PredNet on natural images does not automatically transfer to the artificial dataset, in the first part of this experiment we train PredNet on a subset of the GoL dataset before evaluating it on a different subset of the dataset, as is the normal procedure. Every weight of the network was set randomly before this re-training, so there is no influence of previous experiments.

\begin{figure}[t]
\centering
\includegraphics[width=\linewidth]{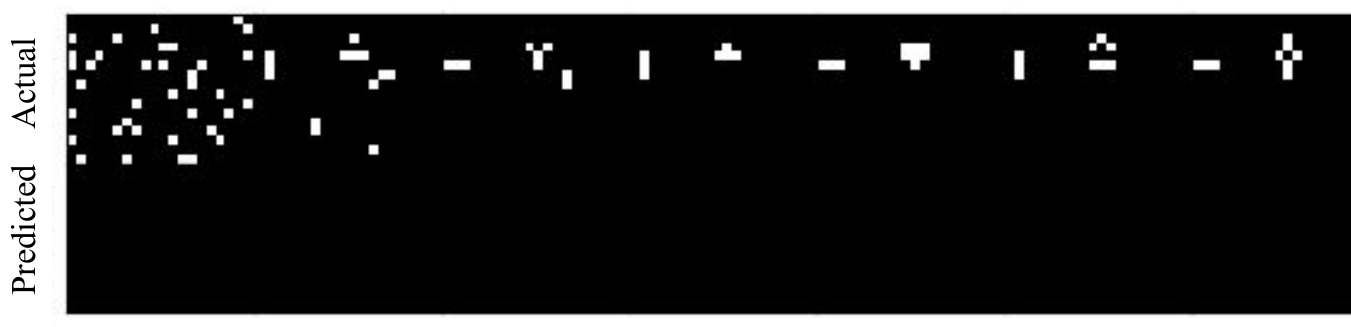}
\caption{{\bf Predictions generated by a PredNet trained and evaluated on random GoL patterns}. PredNet no longer outputs anything as predictions. The rules of the GoL seem not simply impossible to predict using PredNet, but impossible to learn.}
\label{fig:gol_retrained_prednet}
\end{figure}

As shown on Fig.~\ref{fig:gol_retrained_prednet}, the network is unable to learn and performs poorly. This result is unexpected considering that this task is solved by a CNN as shown in the next section, and PredNet has several CNNs with the appropriate filter size of $3~\times~3$ in its architecture.
This suggests that high performance on natural videos comes to the cost of generality for videos exhibiting different types of dynamics (even if these dynamics are totally deterministic and predictable).

\begin{figure}[t]
\centering
\includegraphics[width=\linewidth]{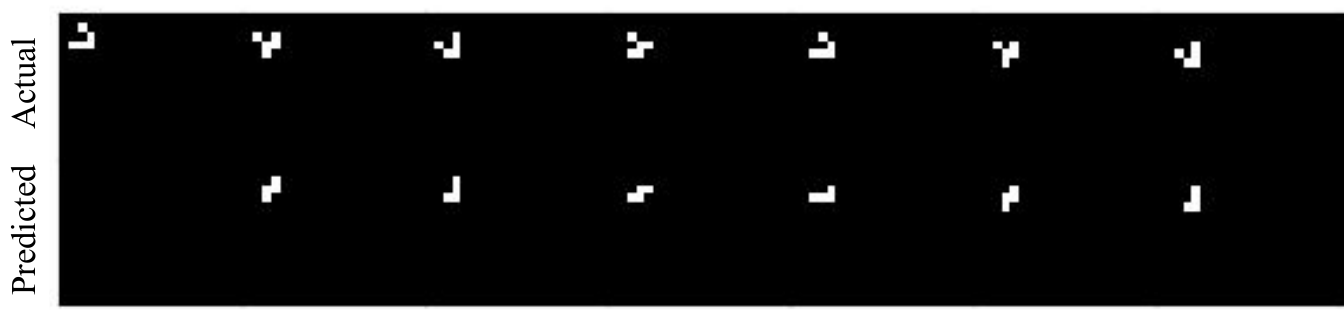}
\caption{{\bf Predictions generated by a PredNet trained only on the exact sequence shown here}. In this case, training data and test data are equal. Even for this simple task, PredNet seems to only be able to imperfectly predict the pattern, with a constant one-cell mistake. In reality, it seems that what was learned was a simple translation of elementary patterns of period 5.}
\label{fig:gol_memory_prednet}
\end{figure}

Finally, we train and evaluate PredNet on a single sequence showing a glider moving from the upper left to the lower right of the frame. In this setting we use the same data for learning and evaluation.
As shown on Fig.~\ref{fig:gol_memory_prednet}, even with these extremely gentle conditions, PredNet is unable to make perfect predictions. What is seems to have learned is a set of periodic translations of period 5 for simple patterns.

\subsection{Simple CNN performance on the GoL}

\begin{figure}[t]
    \includegraphics[width=\linewidth]{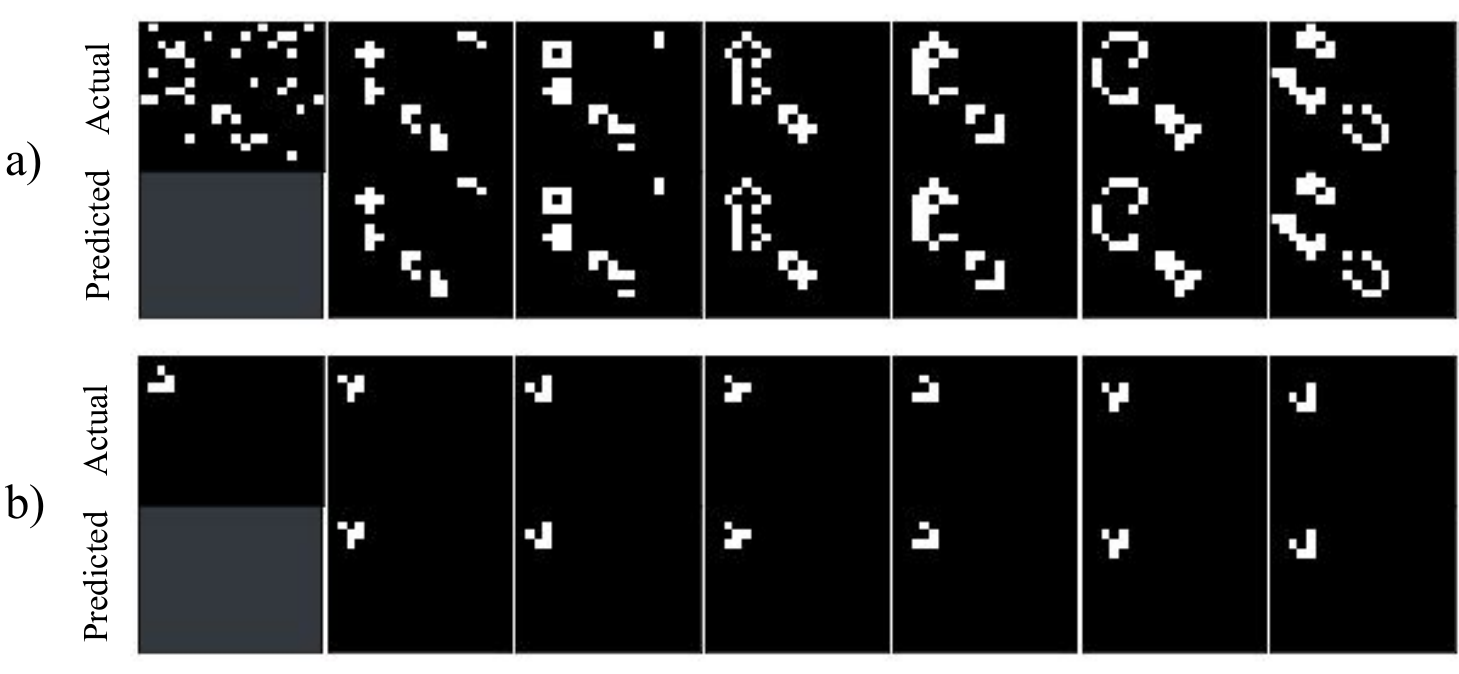}
    \caption{{\bf Predictions generated by the CNN after training on random GoL patterns}.
    (a) Predictions from a random initial state. 
    (b) Predictions for a initial state composed of one glider.
    The performance is close to perfect, an expected result considering that the convolutional filters are the same size as the local neighborhood of the GoL cells.
    }
    \label{fig:gol_cnn}
\end{figure}

We train the CNN on GoL videos generated from initial random patterns.
We then test the CNN on two sequences: (a) a random pattern and (2) a glider pattern where a glider moves from the top left of the scene to the bottom right.
As shown on Fig.~\ref{fig:gol_cnn}, the CNN trained on the random patterns performs almost perfectly on both datasets, as previously reported in~\cite{rappGOL}.
This result is not surprising considering that we fixed the surface of the convolution filters (receptive field of hidden neurons) to 3x3 cells: all the information necessary to predict the next step is contained into every individual filter. In addition, the rules are the same for the whole image, so that rules learned locally can be successfully applied anywhere in the image.

\subsection{Simple CNN performance on KITTI}

\begin{figure}[t]
    \includegraphics[width=\linewidth]{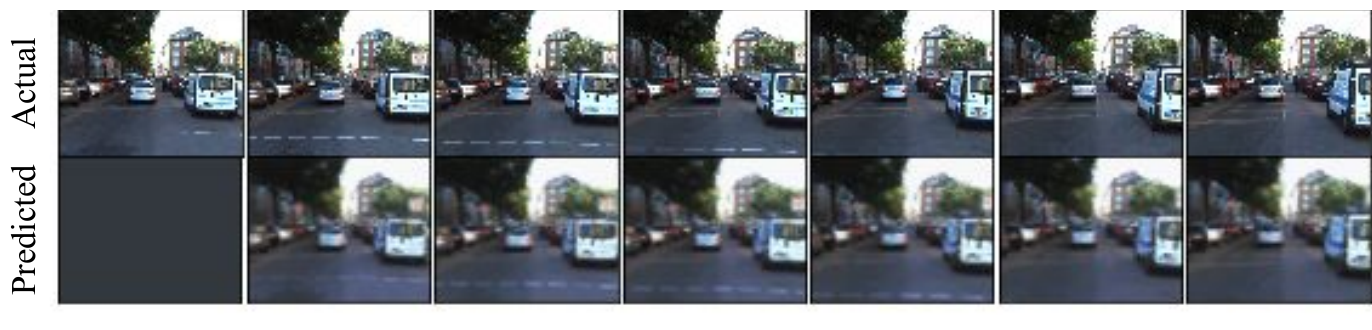}
    \caption{{\bf Predictions generated by the CNN after training on the KITTI dataset}. The performance is poor, especially compared to PredNet (an expected result for natural videos).}
    \label{fig:kitti_cnn}
\end{figure}

As a last experiment, we train the simple CNN on the KITTI dataset. We do not expect good performance, as PredNet was implemented with the sole goal of achieving state of the art performance and simple CNNs are notoriously poor at natural video prediction. The results Fig.~\ref{fig:kitti_cnn} show that indeed, the perfomance is poor. We stopped learning when the number of epochs was 10 because the value of loss did not fall any more. The output of the CNN resembles a fuzzy copy of the last presented frame. As recorded in Table~\ref{tab:mse}, the simple CNN's error is an order of magnitude bigger than PredNet's error, and slightly lower than last frame copy.
This result again shows a trade off, but opposite of PredNet: we have perfect predictions on the artificial dataset and poor predictions on the natural dataset.

\begin{table}[ht]
  \begin{center}
    \caption{Mean squared error (MSE) of next frame predictions.
    As mentioned in the Dataset's subsection, all datasets are set to the same pixel size, so it is meaningful to compare values side by side.}
  \begin{tabular}{l l|c}
    Dataset      & Method               & MSE \\ \hline
      KITTI      & Previous Frame       & 0.0256 \\
                 & PredNet              & 0.0073 \\
                 & Simple CNN           & 0.0164 \\ \hline
    FPSI       & Previous Frame       & 0.0110 \\
                 & PredNet (pretrained) & 0.0059 \\ \hline
    GoL (random) & Previous Frame       & 0.0332 \\
                 & PredNet              & 0.0330 \\
                 & PredNet (re-trained) & 0.0348 \\
                 & Simple CNN           & 0.0002 \\ \hline
    \end{tabular}
    \label{tab:mse}
  \end{center}
\end{table}

\section{Discussion}

We showed that PredNet can generalize without retraining on a completely different natural video dataset. Yet we also found that PredNet cannot learn to predict the dynamics of the GoL. Worse yet, it cannot memorize a short 10-frame sequence of the GoL: a task that should be solvable rote memorization of the mapping between input and output.
This result is surprising since a simple CNN with only one conv-deconv layer did learn the GoL rules, yet PredNet was not able to learn despite internally containing modules that are themselves simple CNNs.

This result shows that there is at least one unsuitable task for the PredNet architecture as it is. In this models, past observations are abstracted and stored internally, and it seems that input images are shifted by an appropriate amount using estimated velocity and rotation information. This is especially striking in the prediction error on frame 3 in Fig.~\ref{fig:pretrained_prednet}-a): the translational motion of the cap that appeared at frame 2 is extrapolated, leading to the prediction of a floating cap at frame 3.
This explains why these models are good for continuous dynamics such as movement of objects in the real world.
On the other hand, in the GoL world, every cell changes in a discrete manner according to the GoL rules. This can be captured by CNNs, but this ability seems to be somehow inhibited by the very architecture that makes PredNet good at shifting and rotating.
Note that we just assumed that PredNet, as current state of the art in video prediction, was a kind of "master algorithm" for natural videos; in reality its failure does not mean that all high-performance predictive algorithms would fail. Yet we cannot help but wonder if a single architecture would be able to perform on both natural and artificial datasets, or if the trade-off in performance cannot be avoided. Supporting this latter hypothesis, the GoL is extremely difficult to predict for human brains; even its own creator, knowing the rules, had to use a physical board with go stones to compute states and still made errors. If there really is a divergence between optimizing predictions for natural and synthetic datasets, then when thinking about an hypothetical general artificial intelligence, we should also consider how to compensate for the counter-intuitive weaknesses that such superior models create for themselves compared to lower-performance models. More generally, it suggests that ``prediction" does not have the same meaning in natural and artificial worlds.

\footnotesize
\bibliographystyle{apalike}
\bibliography{example} 

\newpage
\section{Additional Details}

\subsection{Reversed video prediction}

\begin{figure}[h]
    \includegraphics[width=\linewidth]{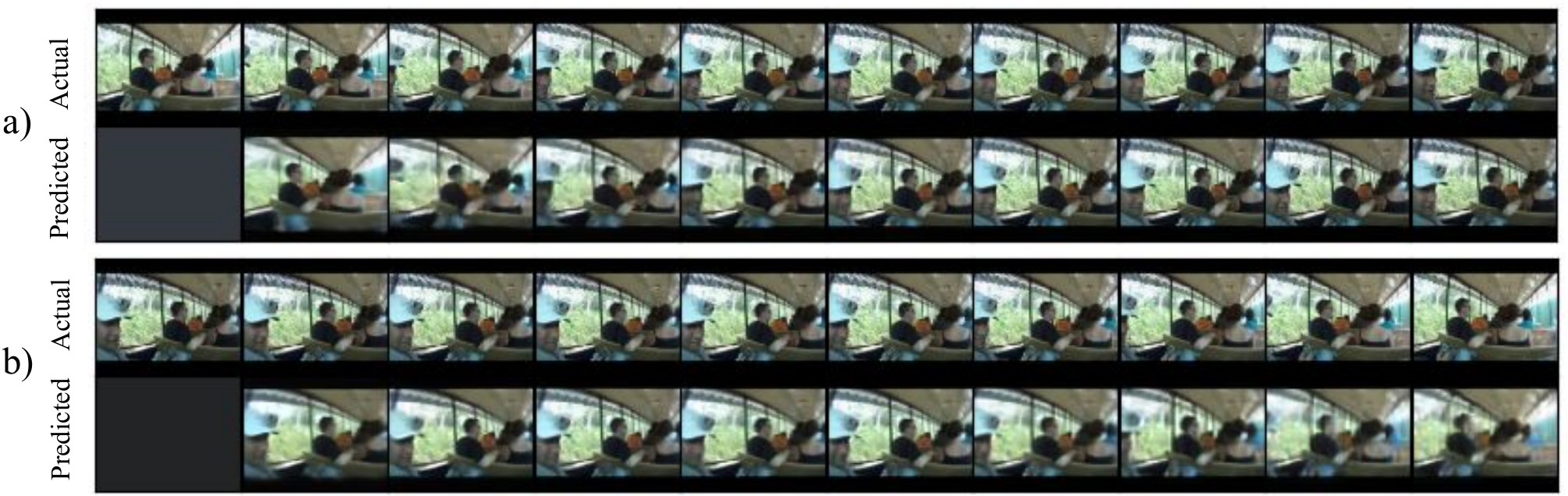}
    \caption{{\bf Predictions generated on the FPSI dataset}.
    (a) Predictions on the regular-time video.
    (b) Predictions on the reverse-time video.
    The predictions are not exactly the same but the performance is similar in both cases.
    }
    \label{fig:reverse_fpsi}
\end{figure}

Although physical laws are time reversible, it is famously difficult for humans to predict videos where the flow of time is reversed. 
We tested whether a PredNet trained on regular-time videos from the KITTI dataset would be able to predict reverse-time videos. The model turned out to be just as good for regular- and reverse-time videos, with a MSE of 0.006 in both cases. Fig~\ref{fig:reverse_fpsi} shows the generated predictions.

\subsection{Simple CNN with smaller filters}

\begin{figure}[h]
    \includegraphics[width=\linewidth]{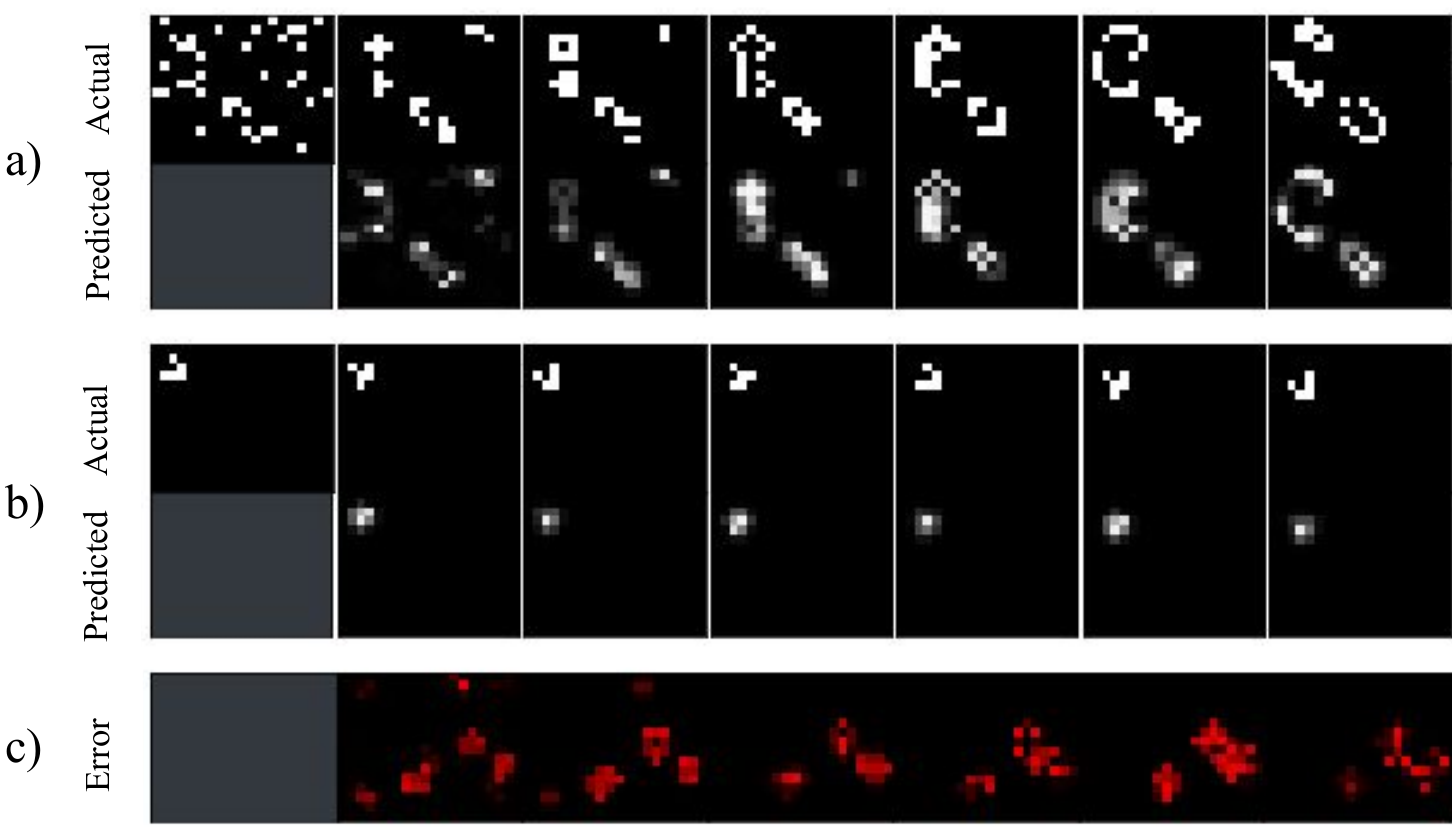}
    \caption{\textbf{Predictions with smaller convolutional filters.}
    The settings are the same as for Fig.~\ref{fig:gol_cnn}, but the filter size is $2~\times~2$ instead of $3~\times~3$.
    (a) Predictions from a random initial state. 
    (b) Predictions for a initial state composed of one glider.
    (c) The difference between actual and predicted images.
    The performance is poor. The filters are too small to capture neighborhood information.
    }
    \label{fig:gol_cnn_kernelsize2}
\end{figure}

When using $2~\times~2$ convolutional filters instead of $3~\times~3$ filters, the performance of the simple CNN model sharply declined as shown in Fig.~\ref{fig:gol_cnn_kernelsize2}. This result shows that the filters are too small to represent the GoL rules. This is not entirely surprising. In the GoL, the next state of a cell is completely determined by the state of eight neighboring cells: all necessary information is contained in a $3~\times~3$ to predict the next state.
When increasing the filter size, the performance was the same as for $3~\times~3$ filters.

\end{document}